\pdfoutput=1

\documentclass[11pt]{article}

\usepackage[]{EMNLP2022}

\usepackage{times}
\usepackage{latexsym}

\usepackage[T1]{fontenc}

\usepackage[utf8]{inputenc}

\usepackage{microtype}

\usepackage{inconsolata}

\usepackage{graphicx}
\usepackage{float}
\usepackage{algorithm}
\usepackage{algpseudocode}

\title{Towards leveraging latent knowledge and Dialogue context for real-world conversational question answering}

 \author{Shaomu Tan *\\ University of Amsterdam, The Netherlands \\ \texttt{s.tan@uva.nl}
         \AND
         Denis Paperno \\ Utrecht University, The Netherlands \\ \texttt{d.paperno@uu.nl}
         }

\begin{document}
\maketitle
\begingroup\def\thefootnote{*}\footnotetext{Work done as an intern at ABN AMRO Bank N.V.}\endgroup
\begin{abstract}
In many real-world scenarios, the absence of external knowledge source like Wikipedia restricts question answering systems to rely on latent internal knowledge in limited dialogue data. In addition, humans often seek answers by asking several questions for more comprehensive information. As the dialog becomes more extensive, machines are challenged to refer to previous conversation rounds to answer questions. In this work, we propose to leverage latent knowledge in existing conversation logs via a neural Retrieval-Reading system, enhanced with a TFIDF-based text summarizer refining lengthy conversational history to alleviate the long context issue. Our experiments show that our Retrieval-Reading system can exploit retrieved background knowledge to generate significantly better answers. The results also indicate that our context summarizer significantly helps both the retriever and the reader by introducing more concise and less noisy contextual information.

\end{abstract}

\section{Introduction}
Answering questions requires an efficient exploitation of knowledge. 
Modern QA systems \cite{karpukhin2020dense, guu2020realm, lewis2020retrieval, izacard2020leveraging} often rely on external knowledge like Wikipedia or databases that contain explicit answers. However, for many real-world QA tasks, such external knowledge either does not exist or can hardly be utilized when facing complex and very personal non-factoid questions. FAQ or How-To documents usually can help users with simple guidance on the basics in real-world applications; yet, there is no guarantee that they can cover a wide range of issues. Under these circumstances, the absence of external knowledge limits the application of QA systems in real-world scenarios. However, conversation logs collected from real conversations conducted between the questioner and the respondent contain valuable information which can be considered as latent background knowledge to answer questions. In this study, we propose a neural Retrieval-Reading system to leverage this latent background knowledge in QA tasks. Our system consists of a retriever and a generative reader. The former retrieves relevant background knowledge to the given question using dialog data, and the reader incorporates the knowledge to answer questions.
\begin{figure}[H]
    \centering
    \includegraphics[width=8cm]{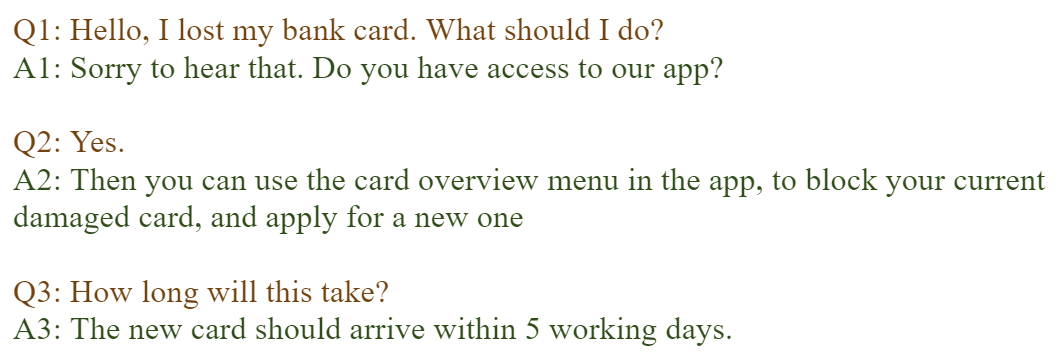}
    \caption{An written test dialogue example in our research. Each conversation turn has a Question ($Q_i$) and an Answer ($A_i$). Dialogue history for $Q_3$ is $\mathrm{\left\{Q_1, A_1; Q_2, A_2 \right\}}$}
    \label{fig:ConvQA-example}
\end{figure}
\noindent Humans seek answers through multiple rounds of dialogue by asking several questions for more comprehensive information. As Figure \ref{fig:ConvQA-example} shows, Conversational Question Answering (ConvQA) tasks require the respondent to consider both the current question and one or multiple specific conversation rounds from the dialog history to infer answers. Previous research \cite{choi2018quac, reddy2019coqa, ohsugi2019simple} has shown that incorporating more dialogue context rounds to a certain degree improves readers' prediction qualities; however, considering too much contextual detail impairs their performance. The reasons are 1.\ readers suffer from processing long sequential information and expect additional computational resources, and 2.\ not all history rounds are helpful for understanding the current question; some may even confuse the machine
. Furthermore, QA data in the real world often tends to be colloquial and informal \cite{faisal2021sd}, covering non-valuable semantic content \cite{ravichander2021noiseqa}, such as greetings and personal and redundant information besides true intent \cite{li2018question}. Such noisy information distracts QA models from capturing the focus of attention. Therefore, we propose a text summarizer based on TFIDF refining contexts to efficiently provide more concise and less noisy contextual information for our Retrieval-Reading system. Our history summarizer can be easily implemented in real-world applications with minimal computing resources in an unsupervised way.\\\\
Our work is conducted on a real industry application scenario, and it makes three important contributions to the field of real-world ConvQA:\\
1. Our research verifies that even without explicit external knowledge, exploitation for latent internal knowledge can enhance the system.\\
2. We find that the history summarizer significantly improves the retriever and reader thus provides a good practical value in real-life ConvQA applications.\\
3. We explored directions of history modelling. Our main finding is that the use of a generative reader shows a promise, as confirmed by quantitative and qualitative evaluations.

\section{Related work}
\textbf{Knowledge Exploitation}
Many modern extractive QA systems \cite{guu2020realm, yang2019end, karpukhin2020dense} provide answer predictions by retrieving documents containing explicit answers and then extract answer spans from them. Nevertheless, in real-world applications, especially in customer service tasks, it is not feasible to extract answer intervals from FAQ documents or conversation data when facing complex and personal non-factual questions. On the other hand, several studies \cite{lewis2020retrieval,longpre2021mkqa,izacard2020leveraging, roller2020recipes} have investigated building Retrieval-Reading systems by augmenting the generative model via using the retrieved external knowledge like Wikipedia documents in order to generate better response predictions. Their work only investigated the external knowledge exploitation and was solely validated on open-domain QA or chatbot domains; we will investigate internal knowledge exploitations using the Retrieval-Reading systems and validate it on a real-world multilingual conversation question answering task.\\\\
\textbf{Context Dependency Issue} 
The introduction of the multi-turn dialogues introduces the dialog history dependency issue to QA systems, it requires machines to process information from a broad dialogue context to understand the current question \cite{zhu2021retrieving, zaib2021conversational,gupta2020conversational}. Two lines of active investigation aim to improve the context dependency issue.  \textbf{1.\ Question rewriting} studies \cite{vakulenko2021question, anantha2020open, chu2020ask} investigate how to rewrite the combination of contexts from dialog history and the current question into a form that machines can better understand. However, these approaches require
rewritten questions as training data for supervised learning. Their implementation also requires large-scale neural network training, which is hardly practical in real-world applications. \textbf{2. History modelling}. Works by \citet{qu2019bert} and \citet{qu2019attentive} have shown that dynamically encoding contextual information from dialog history for the extractive reader result in better answer spans on the QuAC dataset. However, these approaches are limited to answers directly extractable from the context. 
Here, we will extend history modelling to generative models.

\section{Data}
We conducted our research using an industrial multilingual question answering dataset collected from real conversations conducted between bank customer service agents and customers. It consists of Dutch and English and a small amount of other languages such as German. Table \ref{our_dataset_stats} shows statistics of our dataset and compares them with the QuAC\citep{choi2018quac} dataset. Compared to open-domain academic datasets, our dataset is more challenging bacause: 1. Real-world conversations involve a high percentage of non-factoid questions and redundant dialog turns that does not contain users' true intent. 
2. Our data contains a large number of customer support phone numbers, website urls, etc. Most of this information is contained in the actual answers.
3. Longer QA data poses challenges when encoding and selecting contexts from the conversation history. 4. The lack of external explicit knowledge source like Wikipedia makes the task particularly challenging.
\begin{table}[h!]
\centering
\begin{tabular}{lll}
\hline
\textbf{Stats} & \textbf{Our dataset} & \textbf{QuAC} \\
\hline
\# Questions & 339,478 & 90,922 \\
\# Dialogues & 131,725 & 12,567 \\
\# Avg. Tokens in Q & 24.5 & 6.5\\
\# Avg. Tokens in A & 30.0 & 12.6\\
\# Avg. Turn  & 2.6 & 7.2 \\
\hline
\end{tabular}
\caption{\label{our_dataset_stats}
Data statistics for our dataset and the QuAC dataset. \emph{\# Avg. Tokens in Q} and \emph{\# Avg. Tokens in A} present the average token number for all questions and answers. While \emph{\# Avg. Turn} denotes the average number of rounds for all Dialogues.}
\end{table}

\section{Method}
\begin{figure}[h!]
    \centering
    \includegraphics[width=8cm]{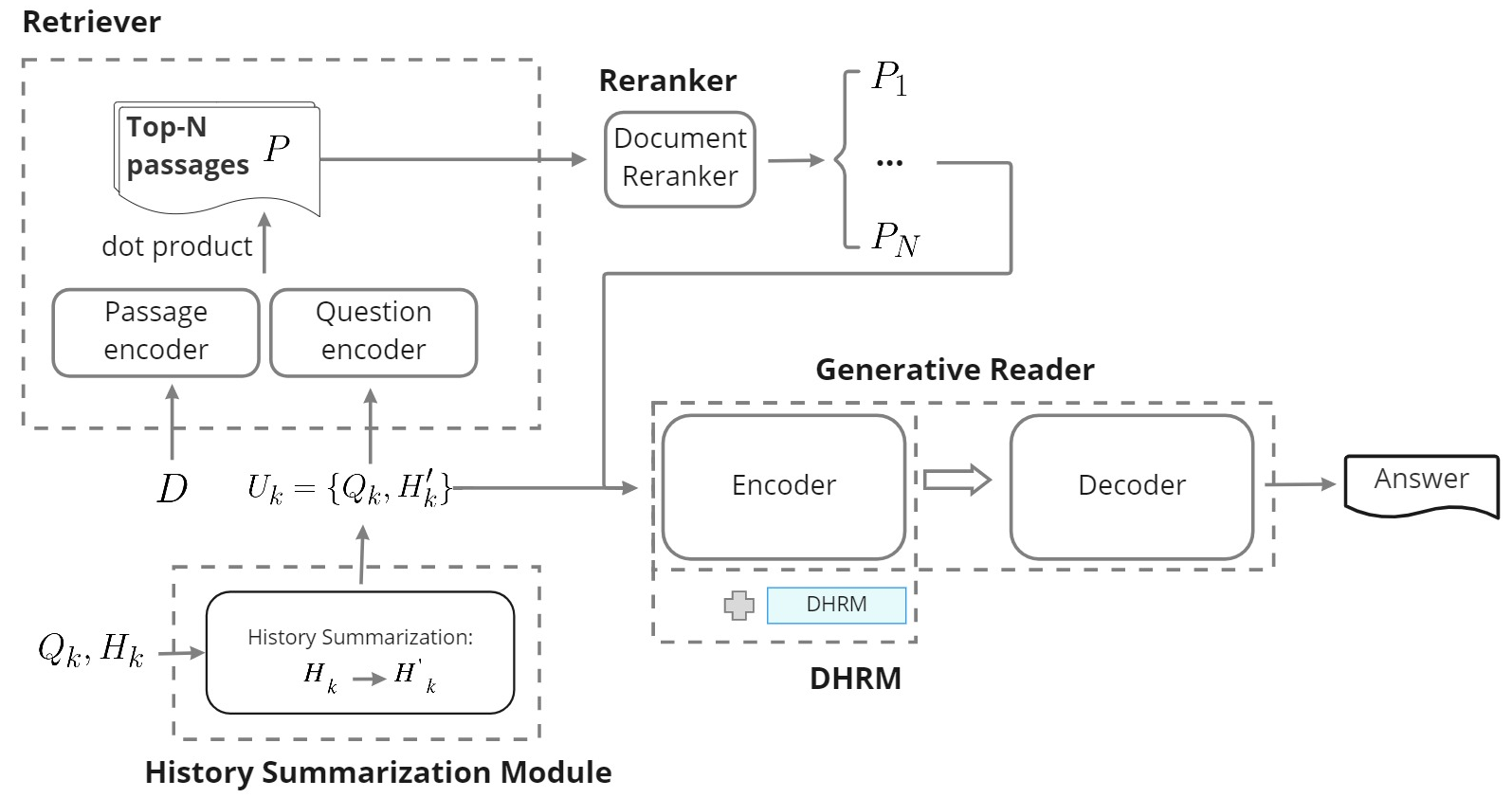}
    \caption{Architecture of the retrieval-reading system in this study}
    \label{fig:QA_system}
\end{figure}

\textbf{Task Definition} Our QA task can be defined as follows: Given a large collection of passages $D$, a query $U_k$ including the current question $Q_k$ and its previous dialog context $H_k$, where the $H_k = \mathrm{\left\{ Q_i, A_i \right\}}_{i=1}^{k-1}$ contains all QA pairs(questions and answers) before the $Q_k$ in this conversation. The goal is to provide an answer $A_k$ for $U_k$ using $D$. In addition, since our data is obtained from real-world customer service, we hardly have the Wikipedia-like articles used in academic data as $D$. Instead, we use answers from QA data as our $D$; hence, in our case, a single element in $D$ is a question answer pair in the dataset.\\\\
\textbf{System Overview} Figure \ref{fig:QA_system} shows the architecture of our QA system. To utilize potential background knowledge in the task of ConvQA, we build a retrieval-reading model that uses a document retriever to provide relevant knowledge and uses a generative reader to predict the answer to the given questions. In addition, we incorporate a reranker, a History Summarization Module (HSM), and a Dynamic History Re-weighting Module (DHRM) in the retrieval-reading architecture to enhance the system. 
We trained all modules on our down-stream task in our experiments, except for HSM which was used without training.
\\\\
\noindent \textbf{Retrieval-Reading System} 
We implement the Retrieval-Reading model to enable the use of 
internal background knowledge in the ConvQA task. High-quality retrieved documents can provide more relevant background knowledge for subsequent generative readers, and ideally answers from top retrieved QA pairs can even be used directly as the predicted answer. 
To validate what model can yield more satisfactory results with real-world data, we consider both sparse (BM25 in \citet{lin2021pyserini}) and dense (DPR in \citet{karpukhin2020dense}) retrievers in our retrieval task. In the retriever, we replace BERT model with Multilingual-BERT (MBERT) in the dual encoder model for efficient multilingual model training. The generative reader in our system makes predictions by considering the query input and background knowledge returned by the retriever. We implement both the mBART\cite{liu2020multilingual} and mT5\cite{xue2020mt5} models as our readers to validate which model will yield better results in our multilingual reading task.\\\\
\textbf{History Summarization Module} We propose a History Summarization Module (HSM) to refine the conversation history context in an attempt to improve both retrieval and reading performance. We implement TFIDF based extractive summarization with stemming because it is commonly used as a strong baseline for summarization tasks. 
Therefore, we choose to keep the head $H_1$ and tail $H_{k-1}$ pairs of QA contexts from the dialog history, and only summarize the content in the middle. This is because, intuitively, the head of a conversation normally contains the user's primary intent, while the tail is most likely to be relevant to the current question since it is the most recent context.\\\\
\textbf{Dynamic History Re-weighting Module} Some history turns are redundant for the current question, while others can help us understand and answer the current question. Thus, our intuition is that by dynamically assigning fewer weights to low-value conversation turns from the dialog history in the reading process, the generative reader could be more capable of handling the ConvQA tasks. Algorithm \ref{alg:Dynamic History Re-weighting Mechanism} shows how Dynamic History Re-weighting Module (DHRM) works with a generative reader. In general, it learns the importance weight of QA pairs from the dialog history, then re-weights those historical turns and tokens in the passages. Compared to work by \citet{qu2019attentive}, our DHRM is different by reweighting both candidate passages and contexts from the conversation history, and our work is intended to implement on the generative reader instead of extractive reader.
\begin{algorithm}[h!]
	\caption{Dynamic History Re-weighting Mechanism (DHRM)}
	\label{alg:Dynamic History Re-weighting Mechanism}
	\begin{algorithmic}[1]
	\State {Input $U_k$: it includes the current question $Q_k$, conversation history $H_k = \mathrm{\left\{ Q_i, A_i \right\}}_{i=1}^{k-1}$, and candidate passages $P_k$.}
	\State {Feed $U_k$ to the encoder to get the contextualized embedding.}
	\State {Utilize mean pooling for the current question embedding and the embedding of each context from the conversation history. 
	This process outputs $[ QS, HS^1, HS^2, ..., HS^{k-1}]$.}
	\State {Use the Bahdanau attention layer to calculate the attention score for each context from the conversation history. After that, pass the attention scores to a softmax layer to compute their attention weights ($ [\alpha^1, \alpha^2, ..., \alpha^{k-1}]$, they are ranging from $[0,1]$).}
	\State {Reweighting tokens in the contexts from the conversation history  by multiplying its corresponding attention weights.}
	\State {Reweighting tokens in the candidate passages. For example, if a token in candidate passage also appeared in $H_2$, then multiply $\alpha^2$ to that the token embedding.}
	\end{algorithmic} 
\end{algorithm}

\noindent \textbf{Passage reranking Module} Inspired by \citet{qu2020open}, we verify whether adding a passage reranker can improve the ranking of the retrieved candidate passages even though the reranker and the dense retriever are based on the same language model without sharing parameters. The reranking strategy can be effective when we cannot obtain satisfactory retrieval ranking results, especially when we cannot take advantage of the large number of background documents being retrieved. We adopted the neural reranker architecture from \citet{nogueira2019passage} and modified it to fit our multilingual scenario by switching to the MBERT.


\section{Experiments}

\textbf{History Contribution Experiment} Firstly, to investigate whether considering complete QA pairs as context yields better results than using the questions or the answers alone, we randomly selected 4000 questions in the test set and then used trained models to predict the ranking of their actual documents among 32,000 documents.\\\\
\textbf{Retrieval QA Experiment} We conduct the complete retrieval experiment using both BM25 and DPR as our retriever to investigate whether relying solely on information retrieval is possible for our QA tasks. Based on the result of history contribution, we select complete  QA pairs as the conversation history context component in this experiment.\\\\
\textbf{Retrieval-Reading Experiment} In this experiment, we implement the retrieval-reading system on our ConvQA task. To further validate that using background knowledge can help the retrieval-reading system to make better predictions than using the generative reader alone, we evaluate a pure generative reader model without retrieved knowledge for comparison. Furthermore, we also compare our model to pure retriever to investigate to what extent the output of information retrieval model can be compared with our retrieval-reading system.\\\\
\textbf{Evaluation metrics}
We measure the average document rank for the experiment of history contribution. It measures the average rank of the true answer to a question among a given large number of documents when the retriever is searching for the answer to that question. For the Retrieval QA Experiment, we utilize top-n retrieval accuracy and three rouge scores \cite{lin2004rouge} as our evaluation metrics. As for the retrieval-reading experiment, in addition to the quantitative evaluation based on rouge scores, we also conducted a qualitative analysis based on a double-blind evaluation by human annotators. In this human evaluation, we compare the scores of three different types of answer candidates on the scales of relevance, correctness, and readability.\\\\
\textbf{Distributed training}
To accelerate the entire training and inference process, we utilized distributed data parallelism and model parallelism to implement uniform distributed training through the Zero Redundancy
Optimizer (ZeRO \citealp{rajbhandari2020zero}) strategy and Pytorch platform \cite{paszke2019pytorch}. 



\section{Results}

\subsection{Results of Retrieval tasks}
As the Table \ref{table:historicalcontextcomposition} shows, both  questions and  answers  are necessary for the retriever, where the latter is more significant than the former. This indicates that considering complete QA pairs on our data helps to better
perform the retrieval task. Our finding is consistent to previous studies conducted on public datasets, such as \citet{reddy2019coqa, zhu2018sdnet}.
\begin{table}[h!]
\centering
\begin{tabular}{lc}
\hline
Models             & Avg. rank \\ \hline
DPR w/Qs             & 170.55    \\ 
DPR w/As             & 158.09    \\ 
\textbf{DPR w/QAs}  & \textbf{156.69}     \\ \hline
\end{tabular}\caption{Evaluation of History Contribution Experiment (Lower is better).}
\label{table:historicalcontextcomposition}
\end{table}
\\
\noindent Table \ref{table: Results-RetrievalTask} shows the results of all our experiments on the machine retrieval task. First of all, it is clear from our results that the dense retriever shows a strong superiority over the sparse retriever, which means that the neural-based retriever can serve as a powerful retriever baseline on real data in industry. Moreover, the introduction of a textual summarizer to refine the context can effectively improve the performance of the retriever and thus offers a good value in real-life applications. Specifically, our approach improves the F1 rouge scores by up to 3.58\% on the retrieval task using TFIDF; hence, in contrast to previous studies on question rewriting \cite{vakulenko2021question, anantha2020open, chu2020ask}, our approach is very easy to implement in the real world application and it does not require the large computational resources that neural models require. Finally, the introduction of a reranker does not achieve better document ranking for retrievers that use neural model-based retrievers, which is not consistent with the conclusions in \citet{qu2020open}.

\begin{table}[h!]

\begin{tabular}{lccc}
\hline
Models & \begin{tabular}[c]{@{}c@{}}Retrieval\\ Accuracy\end{tabular} & Rouge-L Scores \\ \hline
BM25     &  6.00  &  10.23, 14.50, 11.58   \\ \hline
DPR      &  24.75 &  13.67, 20.8, 14.04 \\ \hline
DPR+HSM  & \textbf{26.43}  &  \textbf{18.76, 21.63, 17.62} \\ \hline
\multicolumn{1}{l}{\begin{tabular}[c]{@{}l@{}}DPR+HSM \\ + Reranker \end{tabular}}  & 26.12 & 18.74, 21.06, 17.32 \\ \hline
\end{tabular}\caption{Evaluation of Retrieval Task on test set. Values in the bold font denotes they are better compared to values of other models. For Rouge-L scores, the order of presentation is Precision, Recall, F1 score, and this is consistent with other tables demonstrating rouge scores.}
\label{table: Results-RetrievalTask}
\end{table}

\subsection{Results of Retrieval-Reading tasks}
\begin{figure*}[h!]
    \centering
    \includegraphics[width=14cm]{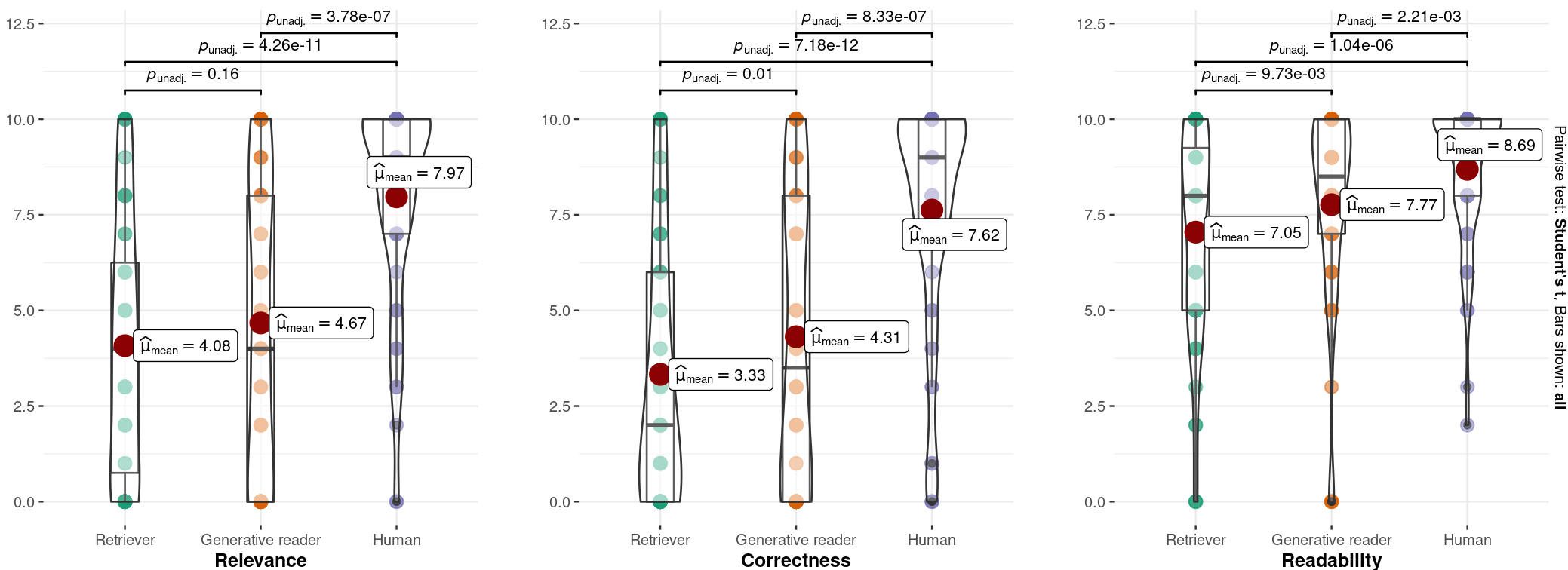}
    \caption{Qualitative Evaluation result of Retrieval-Reading Task on test set. The figure shows the results of experts' ratings on relevance, correctness and readability for three candidate answers. The scores are in the range of [0, 10], higher means better. It also illustrates the pairwise t test result where $p_{unadj}$ represents p-value without adjustment.}
    \label{fig: Qualitative-Retrieval-Reading}
\end{figure*}

Table \ref{table: Results-RetrievalReadingTask2} shows that when we employ only the generative reader without background knowledge, the results can slightly outperform the retriever-only model on all f1 rouge scores. This means that solely using the encoder-decoder model alone without background knowledge can already serve as a strong baseline. Furthermore, our results (in Table \ref{table: Results-RetrievalReadingTask1}) indicate that mT5 performs better than mBart, and our reader can generate better answer predictions when incorporating more background knowledge documents. In addition, our qualitative and quantitative experiments present that the retrieval-reading model can take advantage of the latent background knowledge to make better predictions. Compared to the retriever, our retrieval-reading model improves the precision and F1 of rouge-L scores by up to 6.24\% and 2.87\%. On the other hand, incorporating background knowledge for the reader leads to improvement in all rouge scores, particularly evident in the precision, recall, and f1 metrics for rouge-1 (by 3.57\%, 1.66\%, and 2.11\%, respectively). Figure \ref{fig: Qualitative-Retrieval-Reading} also demonstrate that the retrieval-reading model yields statistically significant improvements than using only the retriever in terms of correctness and readability criteria.
\begin{table}[h!]
\centering
\begin{tabular}{lc}
\hline
Models  & Rouge-L Scores  \\ \hline
\begin{tabular}[c]{@{}l@{}}DPR \\+ mT5 w/5 passages\end{tabular} & 23.49, 20.28, 19.34 \\ \hline
\begin{tabular}[c]{@{}l@{}}DPR\\+ mT5 w/10 passages\\ \end{tabular} & \textbf{25.17}, 21.69, \textbf{20.88} \\ \hline
\begin{tabular}[c]{@{}l@{}}DPR \\+ mBart w/10 passages\end{tabular}& 20.9, \textbf{22.83}, 19.48 \\ \hline
\end{tabular}\caption{Model comparison result of Retrieval-Reading Task on test set. This table compares the performance of the mT5 and mBart models, and compares whether incorporating more background documents helps.}
\label{table: Results-RetrievalReadingTask1}
\end{table}

\begin{table*}[t]
\centering
\begin{tabular}{l|ccc}
\hline
Models                                                                              & Rouge-1 score             & Rouge-2 score           & Rouge-L score             \\ \hline
\begin{tabular}[c]{@{}l@{}}DPR Top-1 answer\end{tabular} & 23.43, \textbf{27.59}, 22.09 & 7.28, 8.84, 7.21  & 18.93, \textbf{22.66}, 18.01 \\ \hline
\begin{tabular}[c]{@{}l@{}}mT5 without background knowledge \end{tabular} & 25.99, 23.58, 22.28 & 8.44, 8.15, 7.55  & 22.1, 20.55, 19.19 \\ \hline
\begin{tabular}[c]{@{}l@{}}DPR + mT5 w/10 passages\\ \end{tabular}& 29.56, 25.24, 24.39 & 10.69, \textbf{10.05}, 9.01 & 25.17, 21.69, 20.88 \\ \hline
\begin{tabular}[c]{@{}l@{}}DPR + mT5 w/10 passages\\ + Summarization\end{tabular}         & \textbf{30.08}, 25.98, \textbf{25.02} & \textbf{11.23}, 9.65, \textbf{9.57} & \textbf{25.61}, 22.35, \textbf{21.44} \\ \hline
\begin{tabular}[c]{@{}l@{}}DPR + mT5 w/10 passages\\ + Summarization\\ + DHRM\end{tabular} & 29.30, 25.83, 24.68 & 10.46, 9.08, 8.94 & 24.79, 22.11, 21.01 \\ \hline
\end{tabular}\caption{Quantitative Evaluation result of Retrieval-Reading Task on test set. The DPR in this experiment inherits the \textbf{DPR+HSM} model from the retrieval experiment. For all rouge scores, the order of presentation is Precision, Recall, F1 score.}
\label{table: Results-RetrievalReadingTask2}
\end{table*}
\noindent Moreover, our History Summarization Module improves the machine's performance on both retrieval and reading tasks in a very efficient way by providing more concise and organized history information. It can be easily deployed on almost any QA system with a retriever or reader, and it is potentially adaptable to most multilingual environments. Finally, we observed that adding the history attention module (DHRM) to the generative model to force the reader to pay less attention to low-value history information did not meet our expectations. This is probably because our data has fewer conversation rounds in general and thus does not have enough data to learn and update parameters for DHRM. In addition, DHRM may also force the generative model to deliberately disregard historical information that is useful for understanding the current question. Therefore, a possible future direction for our study is to observe what questions are more likely to be influenced by DHRM through attention visualization. Our findings are contrary to \cite{qu2019attentive}, suggesting that we have to consider other more efficient history modelling approaches for generative models in future work.
\section{Conclusions}
In this paper,  we present a Retrieval-Reading model with customized modules to explore how to efficiently and effectively perform real-world conversational question answering tasks. We observed that utilizing the retriever to provide relevant internal knowledge to the reader can significantly improve the quality of answer prediction. We also observed that refined contexts bring less noisy information to the retriever and reader, thus allowing them to incorporate more concise and organized information. We hope this work will spur more research on knowledge exploitation and context utilization, which are key elements of how to implement QA and dialogue systems research more efficiently in the real world. In future work, enhancement of document retrieval methods for multilingual, non-factoid data may be a major direction to improve the overall system. And we plan to continue to optimize the efficiency for the conversational scenario leveraging the response of customer feedback from real-world applications.

\section*{Limitations}
One major limitation of our study is that we cannot make the dataset publicly available because of the sensitive nature of the data collected from real customer service conversations. 
In addition, to limit the use of computational resources during processing, we only integrate a small part of retrieved internal knowledge into our system. Expanding retrieved data usage would probably improve our results to a large extent. 

\section*{Ethics Statement}
The data we use in this paper are from real conversations between customers and customer service staff. Therefore, in order to ensure the proper use of data, we strictly comply with the requirements of the GDPR in our research and possible subsequent applications. We use a variety of methods to remove any information that may contain private data during the pre-processing part of the data acquisition and application phase. No information related to real names, customers' email addresses, age or gender, etc.\ was used in this study. In addition, our data is anonymized at the time of acquisition to ensure that no other user identifying information is retained, such as IP addresses.

\bibliography{anthology}
\bibliographystyle{acl_natbib}

\newpage
\appendix

\section{Appendices}\label{sec:appendix}
\textbf{Questionnaire examples of retrieval-reading experiment}\\
To further validate whether a retrieval-reading model that considers background knowledge can generate better responses than a retriever, we invited internal experts from the company to conduct a questionnaire-based double-blinded experiment. In the experiment, a total of 64 questions (32 in English and 32 in Dutch) and their corresponding historical contexts were randomly selected from the test set as examples for the questionnaire. Below are some of the examples in the questionnaire. For faster viewing, we have bolded the core parts of questions and the reasonable answer fragments. Note that we have not bolded any fragments in the real questionnaire.\\\\

\subsubsection{English Examples}

\begin{figure}[H]
    \centering
    \includegraphics[width=8cm]{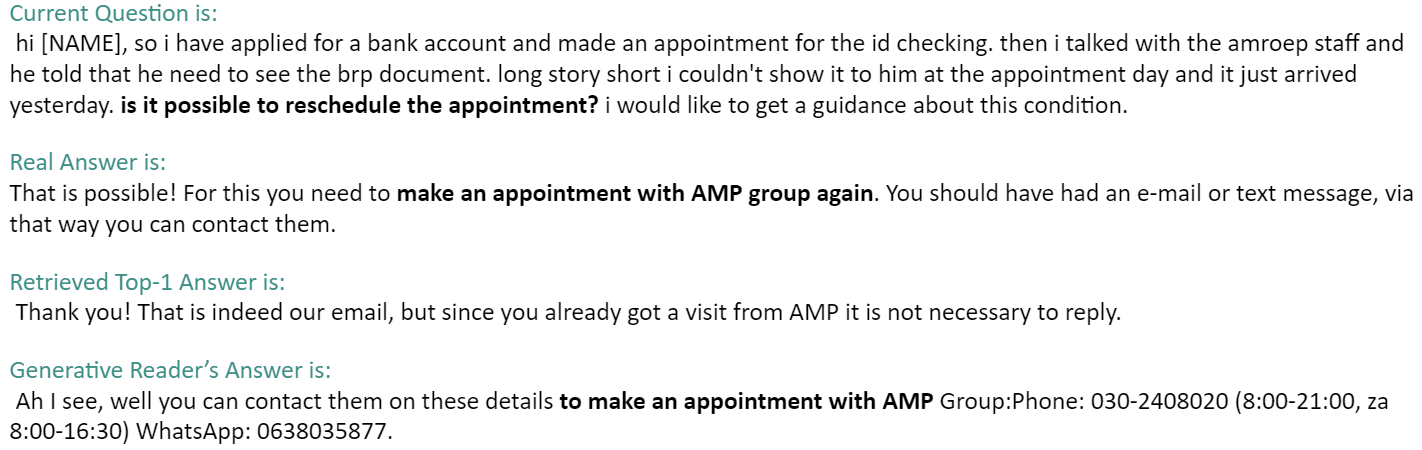}
    \caption{English example 1}
    \label{fig:Example_en_01}
\end{figure}

\begin{figure}[H]
    \centering
    \includegraphics[width=8cm]{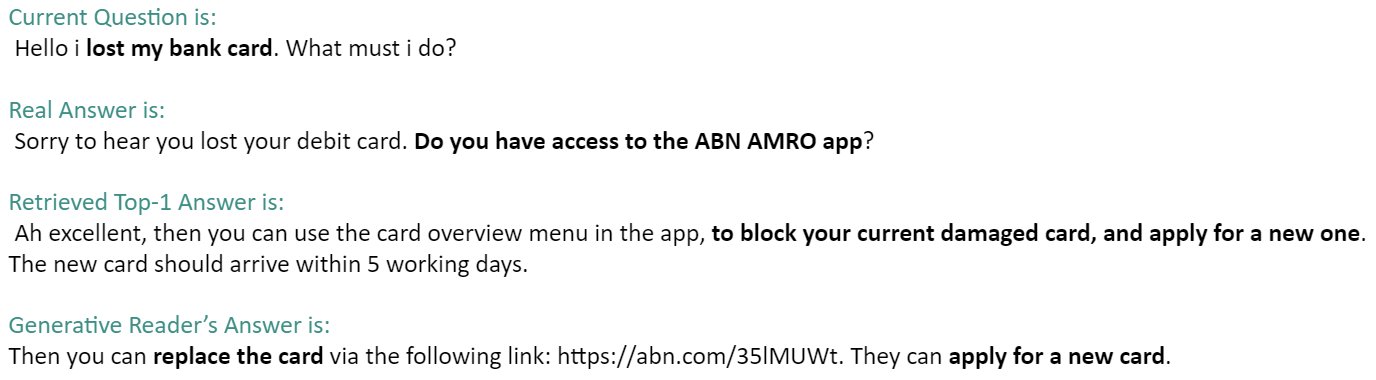}
    \caption{English example 2}
    \label{fig:Example_en_02}
\end{figure}

\subsubsection{Dutch Examples}

\begin{figure}[H]
    \centering
    \includegraphics[width=8cm]{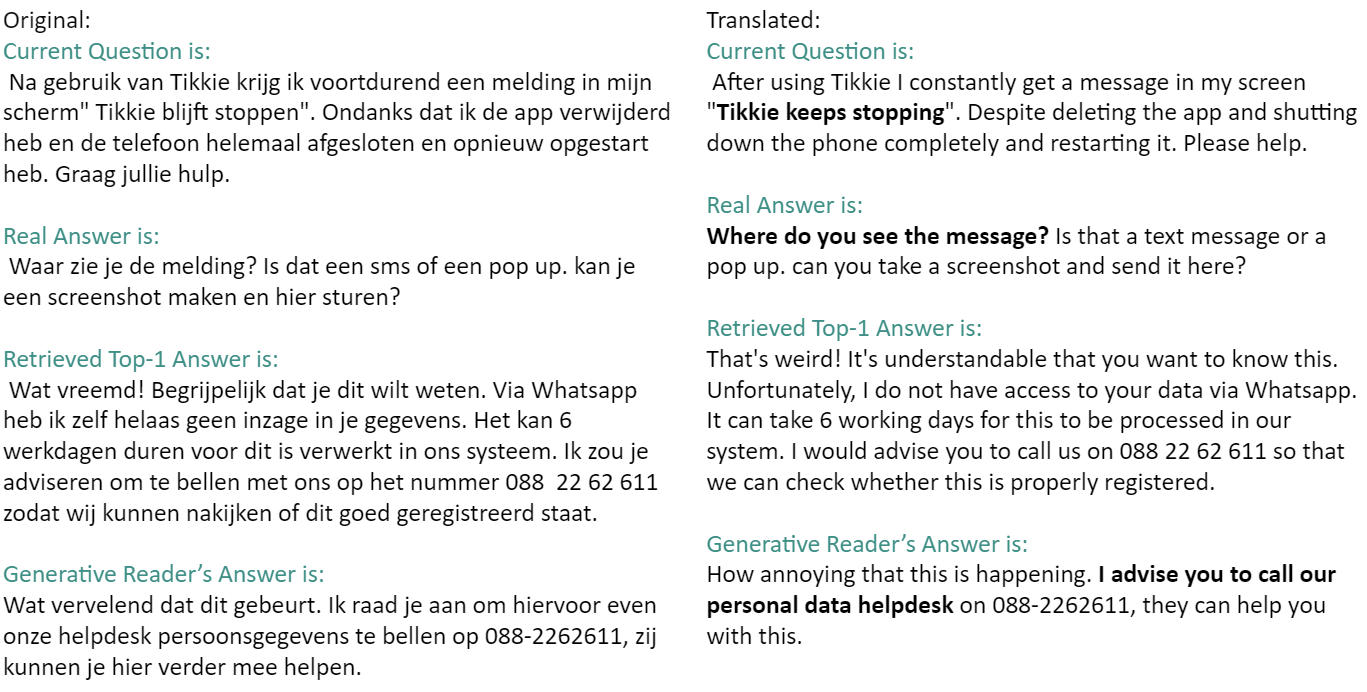}
    \caption{Dutch example 1}
    \label{fig:Example_nl_01}
\end{figure}

\begin{figure}[H]
    \centering
    \includegraphics[width=8cm]{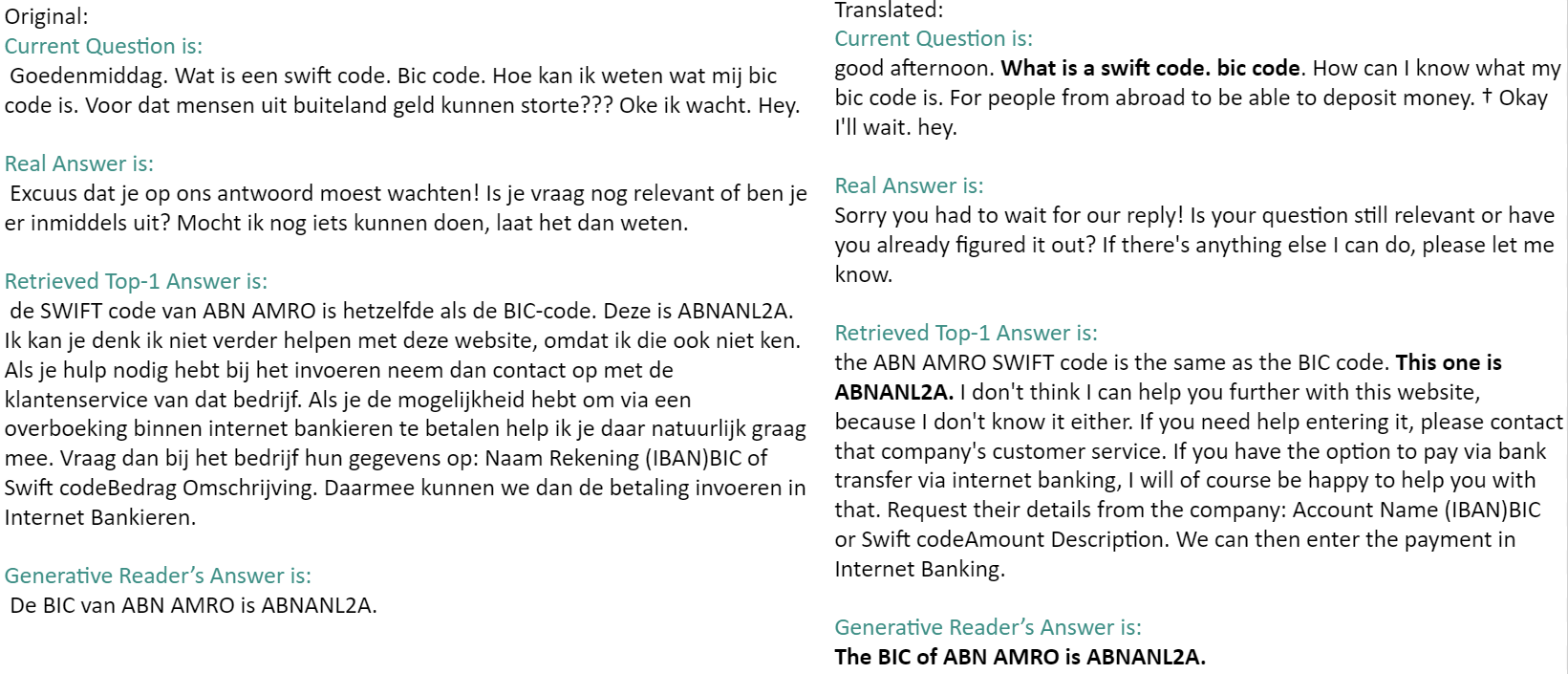}
    \caption{Dutch example 2}
    \label{fig:Example_nl_02}
\end{figure}

\begin{figure}[H]
    \centering
    \includegraphics[width=8cm]{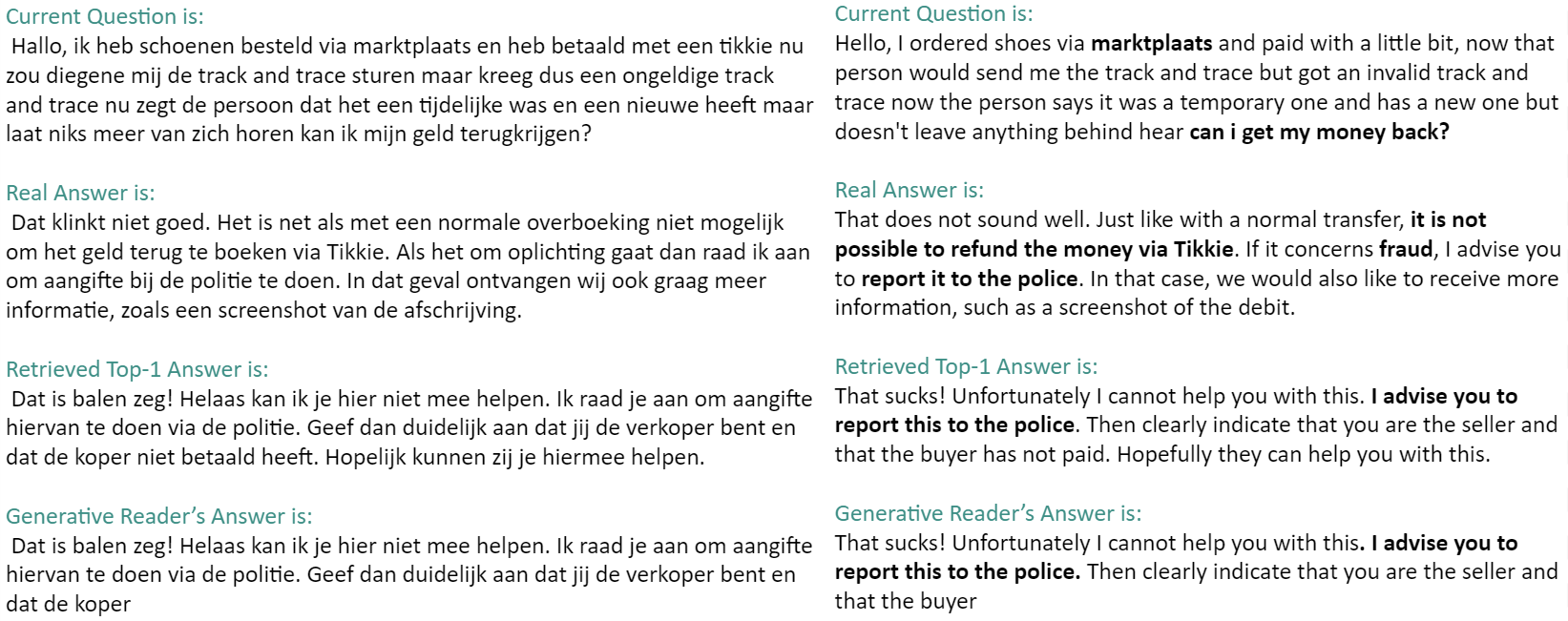}
    \caption{Dutch example 3}
    \label{fig:Example_nl_03}
\end{figure}

\end{document}